\documentclass[journal]{IEEEtran}

\IEEEoverridecommandlockouts                              % This command is only needed if 
\usepackage{amsmath}
\usepackage{multirow}
\usepackage{amssymb}
\usepackage{epsfig} % for postscript graphics files
\usepackage{amsmath} % assumes amsmath package installed
\usepackage[ruled,linesnumbered]{algorithm2e}
\usepackage{graphicx}
\usepackage{caption}

\usepackage[flushleft]{threeparttable}%加表格描述用的

\usepackage{tabularx} % 使用tabularx包来灵活控制表格宽度

\usepackage{color}
\usepackage{cite}

\usepackage{textcomp}
\usepackage{stfloats}
\usepackage{url}
\usepackage{verbatim}
\usepackage{graphicx}
\usepackage{cleveref}
\usepackage{algorithmic}

\usepackage[utf8]{inputenc}

\usepackage[english]{babel}

\usepackage{float}

\addto\captionsenglish{}
\usepackage{subfigure}
\title{\LARGE \bf
Attention and Risk-Aware Decision Framework for Safe Autonomous Driving
}

\author{Zhen Tian$^{1}$, %~\IEEEmembership{Member,~IEEE,}
 Fujiang Yuan$^{2}$, Yangfan He$^{3}$, Qinghao Li$^{4}$, Changlin Chen$^{5}$, Huilin Chen$^{6}$, Tianxiang Xu$^{7}$, Jianyu Duan$^{8}$, Yanhong Peng$^{2,*}$, and Zhihao Lin$^{1,*}$% <-this % stops a space
\thanks{$^{1}$School of Engineering, University of Glasgow, Glasgow, G12 8QQ, U.K.}%
\thanks{$^{2}$College of Mechanical Engineering, Chongqing University of Technology, Chongqing, 400054, China.}%
\thanks{$^{3}$ Computer science with the University of Minnesota - Twin Cities, Minneapolis, MN, USA.}%
\thanks{$^{4}$Department of Computer Science, University of Liverpool, Liverpool L69 3GJ, United Kingdom.}%
\thanks{$^{5}$School of Instrument
Science and Optoelectronics Engineering, Hefei University of Technology,
Hefei 230009, China.}%
\thanks{$^{6}$Faculty of Computer Science and Information Technology, University of Malaya, Kuala Lumpur, Malaysia.}
\thanks{$^{7}$ School of Software and Microelectronics, Peking University, Peking, China.}
\thanks{$^{8}$ School of Transportation Science and Engineering, Beihang University, Beijing 100191, China.}

}%

\begin{document}

\maketitle
\pagestyle{empty}  % no page number for the second and the later pages
\thispagestyle{empty} % no page number for the first page

%%%%%%%%%%%%%%%%%%%%%%%%%%%%%%%%%%%%%%%%%%%%%%%%%%%%%%%%%%%%%%%%%%%%%%%%%%%%%%%%
%Existing autonomous driving systems for the ramp scenario do not consider factors such as the road complexity, complex driving styles of surrounding human-driven vehicles (HDVs), and the lane-changing intention of other vehicles. 
%The proposed method consists of three main components: a driving style grading module, a module for quantifying the prediction errors, and an intelligent decision module. The driving style grading module assigns a grade to each vehicle based on its driving safety features, inverse time to collision and time headway. The module for quantification of prediction errors reduces the decision errors caused by the prediction stage. The intelligent decision module uses error ellipse and game theory to select a suitable time point for entering or exiting the ramp that maximizes the utility function of the AVs while considering the driving styles of surrounding HDVs. 
\begin{abstract}
Autonomous driving has attracted great interest due to its potential capability in full-unsupervised driving. Model-based and learning-based methods are widely used in autonomous driving. Model-based methods rely on pre-defined models of the environment and may struggle with unforeseen events. Proximal policy optimization (PPO), an advanced learning-based method, can adapt to the above limits by learning from interactions with the environment. However, existing PPO faces challenges with poor training results, and low training efficiency in long sequences. Moreover, the poor training results are equivalent to collisions in driving tasks. To solve these issues, this paper develops an improved PPO by introducing the risk-aware mechanism, a risk-attention decision network, a balanced reward function, and a safety-assisted mechanism. The risk-aware mechanism focuses on highlighting areas with potential collisions, facilitating safe-driving learning of the PPO. The balanced reward function adjusts rewards based on the number of surrounding vehicles, promoting efficient exploration of the control strategy during training. Additionally, the risk-attention network enhances the PPO to hold channel and spatial attention for the high-risk areas of input images. Moreover, the safety-assisted mechanism supervises and prevents the actions with risks of collisions during the lane keeping and lane changing. Simulation results on a physical engine demonstrate that the proposed algorithm outperforms benchmark algorithms in collision avoidance, achieving higher peak reward with less training time, and shorter driving time remaining on the risky areas among multiple testing traffic flow scenarios.

\end{abstract}

\begin{IEEEkeywords}
Autonomous driving, driving styles, interactive driving, lane changing, uncertainty quantification.
\end{IEEEkeywords}

%%%%%%%%%%%%%%%%%%%%%%%%%%%%%%%%%%%%%%%%%%%%%%%%%%%%%%%%%%%%%%%%%%%%%%%%%%%%%%%%
\section{INTRODUCTION}

\IEEEPARstart{A}{U}{T}{O}{N}{O}{M}{O}{U}{S} driving with human driving vehicles (HDVs) is a challenging problem that requires reliable and interactive decision making because of complex intentions of HDVs~\cite{lin2025multi,lin2025safety,lin2025safetyr,lin2024enhanced}. The driving scenarios are complex with multiple lanes and HDVs. The intentions of HDVs mainly include lane keeping, lane changing, acceleration, and deceleration, which makes safe driving more difficult~\cite{li2025efficient,li2025adaptive}. Therefore, safety-orientated decision making dominates the autonomous driving because unexpected HDVs' actions are often encountered~\cite{tian2025risk}. To minimize the effects of these unexpected HDV's actions, two model-based approaches have been developed. The first approach aims to model the intentions of HDVs, and the second approach is to model the possible interactive decisions by game theory~\cite{11041353,zheng2025enhanced,zheng2025mean}. Despite the relative high precision, The first approach is limited by its reliance on rule-based classification. For instance, certain algorithms predict lane-changing intentions when the lateral distance varies consistently for three seconds. However, in real situation, vehicles may complete a lane change within this time frame, rendering the rule-based classification imprecise in some scenarios. On the other hand, the second method's limitation stems from excessive cautious decisions based on a worst-case scenario assumption. In reality, not all situations are the worst, but decisions made for the worst-case can negatively impact other factors, like velocity. Therefore, the model-based methods cannot be effectively applied into the interactive driving with HDVs.
\begin{figure*}[ht]%[h]%[!p]
    \centering
    \includegraphics[width=0.9\linewidth]{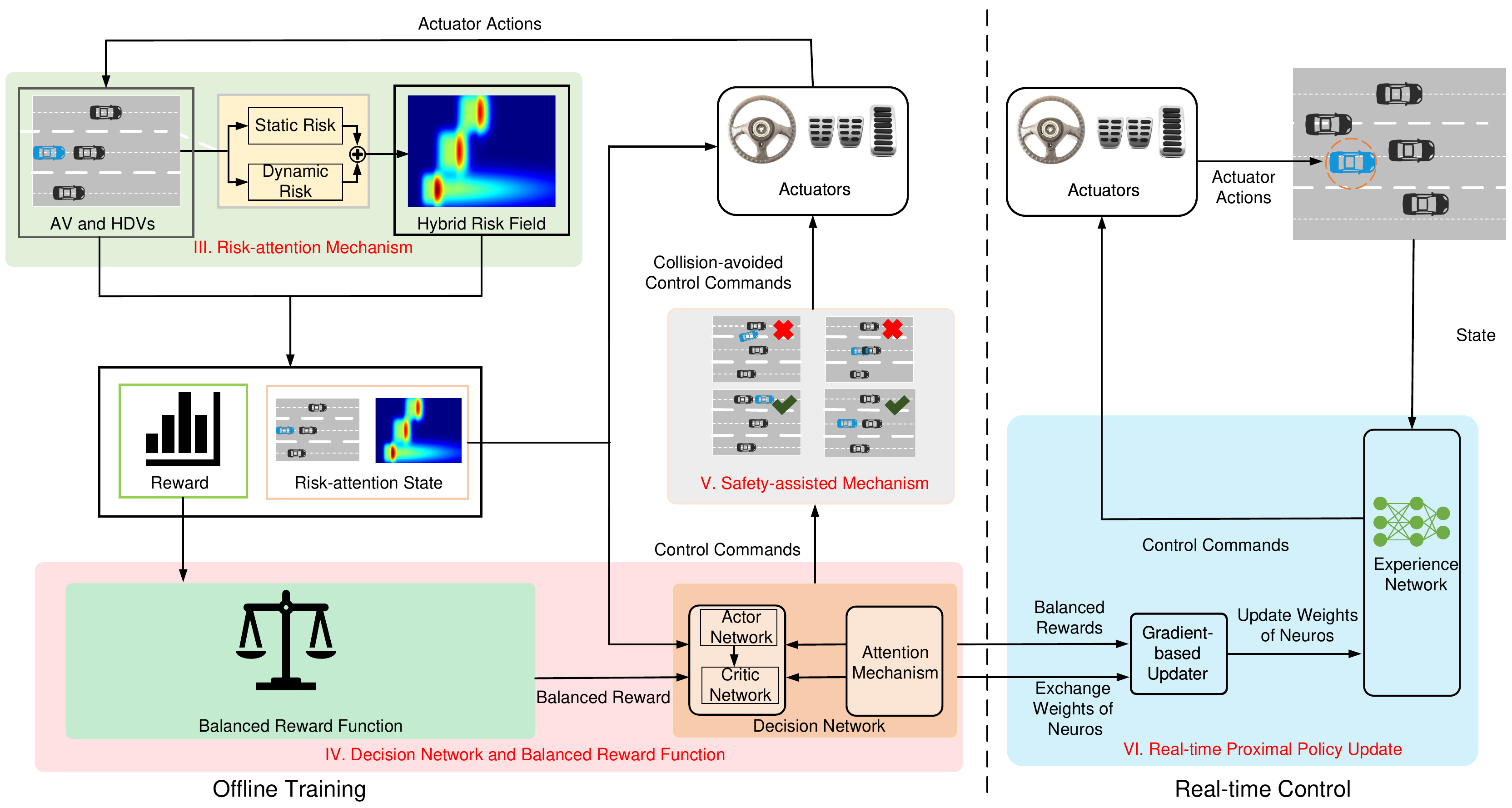}
    \caption{Diagram of the autonomous driving algorithm using the RIBPPO-S.}
    \label{fig2}
\end{figure*}
To address the limitations aforementioned, two methods are proposed. The first method quantifies the risk levels surrounding HDVs. Compared to model-based method, the risk-quantified method directly represents the risk level, eliminating the need to predict HDV intentions or make overly cautious decisions. The risk interaction between Autonomous Vehicles (AVs) and HDVs is evaluated using various states, such as the velocity and relative distance, allowing for the identification of a dependable area to avoid risk. A typical risk-quantified method is the artificial potential field (APF)  \cite{triharminto2016novel}. By utilizing the attractive and repulsive force fields on the target lane and surrounding HDVs, APF guides the AV to the targetlane. A framework that combines the APF with reinforcement learning is proposed in \cite{yao2020path},  achieving collision avoidance with dense obstacles. The primary issue with APF lies in its rigid application of forces. APF treats all areas around the vehicle with the same level of risk, but APF ignores the fact that the front of a car faces more danger than other parts. Moreover, APF is not able to reflect the future tendency of movable objects. Therefore, APF fails to precisely define a separate and predictable risk-avoidance framework for the front vehicles (FV) and rear vehicles (RV) during the interactive driving.

The second method empowers AVs to engage fully with HDVs, formulating safe decision-making strategies through extensive interaction. Learning-based methods facilitate the decision-making exploration by allowing complete interaction with the environment. Reinforcement learning (RL), as a basic learning-based method, is a powerful technique that learns optimal decisions in dynamic scenarios~\cite{lu2023event,wu2022uncertainty}. Therefore, RL is capable to adapt to the the ever-changing nature of interactive driving. Deep reinforcement learning (DRL), as an development of RL by leveraging deep neural networks to approximate complex functions, enables agents to learn from high-dimensional inputs for more complex decision-making. Additionally, \cite{yeom2022deep} achieved the collision-free path planning of mobile robots, surrounded by various obstacles, through the application of DRL. It demonstrated the outstanding capability of DRL to enhance both the safety in autonomous driving. Furthermore, \cite{muzahid2022deep} proposes a DRL powered driving system to effectively avoid collisions when faced with sudden changes in the environment.


Existing DRL algorithms, such as proximal policy optimization (PPO), perform well in short-term driving scenarios. However, these algorithms still encounter challenges in the learning during the long-duration driving. For example, it is difficult for PPO to obtain the global optimal solution in long sequences. Therefore, the improper decisions may lead collisions. Furthermore, the feature map cached by the decision network is crucial for understanding the risks of collisions. To this end, a risk-attention, image-efficient, and balanced reward-orientated PPO with safety-assisted mechanism (RIBPPO-S) is proposed in this paper. The risk-attention mechanism uses a hybrid risk field evaluates the risk levels between AV and HDVs, offering a quantified risks area for collision avoidance. The inputs of the decision network are the local images, combining with the risk-attention field. To read key information from the images, an image-efficient and attention-orientated decision network is proposed. The decision network processes key sections of images and generate collision-attention-based feature maps for safe control commands. The safety-assisted mechanism monitors the actions that pose collision risks during lane maintenance and lane changing maneuvers. Furthermore, to optimize the future steps, a balanced reward function is proposed to consider both the historical and the prospective actions. 

The improved PPO with the curiosity mechanism is illustrated in Fig. \ref{fig2}. The whole algorithm is composed of offline training and real-time control. The decision network produces control commands in offline training, while the experience network generates real-time control commands. In offline training, the decision network learns to generate control commands based on both actions and curiosity rewards. A gradient-based control policy updater is introduced to the experience network, aimed at generating a safe and collision-free strategy. The main contributions of this paper include
\begin{itemize}
    \item  The collisions are avoided by designing an safety-assisted mechanism and image-efficient decision network with local perception. The image-efficient decision network improves the capability of processing key sections of the raw images compared to traditional actor-critic network. 
   \item The average velocity from the starting point to the target point is significantly increased, and the time driving on the risky area is significantly decreased. The training efficiency of racing sequence is improved by the RIBPPO-S over benchmark algorithms.
   \item The collision-free and high-speed driving in various traffic flows are achieved by the proposed balanced reward function. The challenge of maintaining balanced exploration among long sequences is solved by the balanced reward function.
  \end{itemize}

The rest of the paper is organized as follows: Section II summarizes the related works. Section III introduces the risk-attention mechanism. Section IV presents the decision network including the network structure and the control policy update process. Section IV describes the details of curiosity-assisted training optimization. Section V illustrates the balanced reward function. Section VI presents the safety-assisted mechanism. Section VII elaborates the real-time proximal policy update mechanism. Section VIII demonstrates the simulation results. Section IX presents the discussions. Section X draws the conclusions.
\section{Related Works}
State-of-the-art results of using DRL have been demonstrated in autonomous vehicles~\cite{lyu2022advance,liu2021methodology,khalil2022exploiting,li2022pomdp}. Recently, a set of DRL algorithms with exceptional performance have attracted interest, such as Q-Learning~\cite{xu2020nash}, deep deterministic policy gradient (DDPG)~\cite{yang2023towards}, soft actor-critic (SAC)~\cite{haarnoja2018soft}, and PPO algorithms~\cite{9693175}.

In Q-Learning, the state-action value function is utilized to determine the best action in a given state~\cite{xu2020nash}. Correct actions are selected by Q-Learning in autonomous driving despite numerous safety constraints~\cite{min2018deep}. For instance, a combined RL was proposed for lane changes on straight highways in~\cite{min2018deep} using deep Q-Learning networks. However, only simple tasks and scenarios were considered and tested in~\cite{min2018deep}, without considering more complex tasks. Moreover, another drawback of Q-Learning is the low training efficiency~\cite{gu2016continuous}. 

DDPG uses deep neural networks to approximate the control policy~\cite{yang2023towards}. With the suitability for processing high-dimensional data, various demonstrations of using DDPG have been given in autonomous driving~\cite{basile2022ddpg}. For example, a DDPG algorithm was proposed for safety driving within an end-to-end architecture~\cite{basile2022ddpg}. Improved DDPG models have been proposed for improved training efficiency and results~\cite{dong2020mobile}~\cite{hebaish2022towards}. However, the key issue is that DDPG holds an absolute result from the control policy. The absolute result prevents the exploration of more possible actions and limits the adaptability in diverse driving scenarios. 

SAC employs a soft policy, which maximizes a soft value function rather than directly maximizing the expected reward. The soft policy encourages extensive explorations of possible actions. A collision-free car following is achieved by using the SAC, without considering other autonomous driving behaviours~\cite{wang2022velocity}. However, SAC holds a slower training speed, and lower peak reward over PPO for the task of collision avoidance among multiple AVs~\cite{muzahid2022deep}.

PPO utilizes the control policy in a probability distribution and offers faster exploration improvements over DDPG~\cite{10309299}. PPO has been used to create control models for multi-agent driving scenarios~\cite{wei2019mixed}. PPO has been developed to generate smart driving strategies that balance safety and efficiency for crowded highway traffic~\cite{ye2020automated}. In PPO, the reward function is often connected to the averaged performance. The averaged rewards are comprised of unimportant steps in some cases, rather than focusing on the optimization in key steps. Therefore, the training efficiency of PPO for long-length training sequences is low.

In summary, DRL algorithms encounter challenges in fully exploring the environment, often relying on vast interactions with the environment and exhibiting lower training efficiency in long sequences. The PPO addresses some of these challenges by employing probability distributions for exploration and removing reliance on absolute results. However, the training efficiency of PPO diminishes in long sequence tasks like driving on long-length highway. Moreover, the inherent risk of collisions with HDVs during complex interactions remains unresolved due to its averaged attention given to the input features. Furthermore, balanced exploration is essential in long driving sequences to effectively construct the probability distribution of PPO. Additionally, numerous detectable collision scenarios  extend training duration, decreasing the training efficiency. To mitigate these issues, a risk-attention, image-efficient, and balanced reward-orientated PPO with safety-assisted mechanism is proposed in this paper. The proposed risk attention mechanism quantifies the risk levels around surrounding HDVs, thereby decreasing the likelihood of AV remaining in hazardous zones. The image-efficient mechanism integrates an attention network within the decision network to focus on high-risk areas. Furthermore, the balanced reward function facilitates balanced exploration from a global perspective. Additionally, the safety-assisted mechanism filters out certain predictive and detective collisions, reducing the amount of training scenarios and improving the training efficiency. As a result, the challenges of training and collision avoidance of DRL in long sequences are addressed by introducing the risk-attention mechanism, the image-efficient mechanism, the balanced reward function and the safety-assisted mechanism.
\begin{figure*}[ht]%[h]%[!p]
    \centering
    \includegraphics[width=0.9\linewidth]{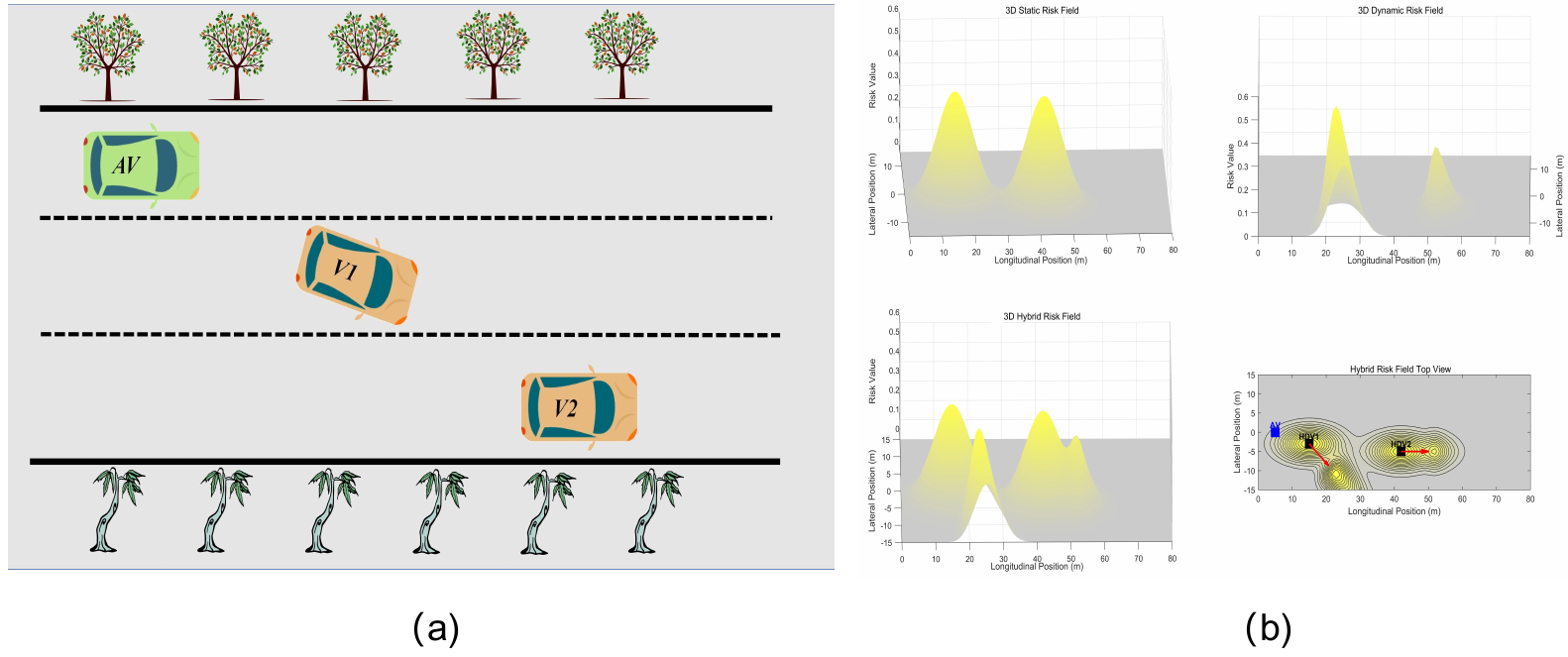}
    \caption{Structure of the image-efficient actor-critic network.}
    \label{fig11}
\end{figure*}
\section{Risk-attention Mechanism}
%\subsubsection{The driving scenario}
\subsection{Hybrid Risk Filed}
This paper proposes a risk-attention mechanism utilizes hybrid risk field to measure the risk levels of HDVs. The hybrid risk field consists of static and dynamic risk fields. The static risk field is a field with different levels of risks evaluated by the static conditions of the obstacle vehicles. The dynamic risk field is evaluated by the relative dynamic properties between the AV and the HDVs.
\subsubsection{Static Risk Filed}
The static risk field is used to generate an static range of risks around the HDVs to quantifiy the risk levels. For a range of HDVs, the following equations are formulated
$$ r_{s}(t)=(\frac{(x_\textrm{{AV}}(t)-x_\textrm{{HDV}}(t))^2 }{\xi _{x}^2} )^\rho +(\frac{(y_\textrm{{AV}}(t)-y_\textrm{{HDV}}(t))^2 }{\xi _{y}^2} )^\rho $$
\begin{equation}
 R_{s}(t)=\varepsilon _\textrm{{obs}}e^{-r_{s}(t)}   
\end{equation}
where $r_{s}(t)$\, is a coefficient of static risks at time $t$\,, $x_\textrm{{AV}}(t)$\, and $y_\textrm{{HV}}(t)$\, are the longitudinal and lateral coordinate of HV at time $t$\, respectively, $x_{obs}(t)$\, and $y_{obs}(t)$\, are the longitudinal and lateral coordinate of each obstacle vehicle at time $t$\, respectively, $\rho$\, is a coefficient to quantify the level of risks, $R_{s}(t)$\, is the magnitude of static risks, $\varepsilon$\, is a weight factor to quantify the risk level of obstacle vehicles, $\xi _{x}$\, and $\xi _{y}$\, are defined with error variance from the prediction stage by error ellipse. The risk value depends on the distance between the host vehicle and the obstacle vehicles, which provides a reference for selecting the sampling points.
\subsubsection{Dynamic Risk Filed}
The dynamic risk field is used to generate a risk range that measures the relative kinematic relationships between the host vehicle and the obstacle vehicles. The dynamic risk field takes into account the velocity difference between the vehicles, which enhances the reliability of the risk range. We establish the dynamic risk field as follows
$$ r_{d}(t)=(\frac{(x_\textrm{{AV}}(t)-x_\textrm{{HDV}}(t))^2 }{\xi _{v}^2} )^\lambda   +(\frac{(y_\textrm{{AV}}(t)-y_\textrm{{HDV}}(t))^2 }{\xi _{y}^2} )^\lambda $$
\begin{equation}
 v_\textrm{{relative}}\left\{\begin{matrix} =1 & if\,\,v_\textrm{{HDV}}>v_\textrm{{AV}}\\ =-1 &if\,\,v_\textrm{{HDV}}\le v_\textrm{{AV}}\\\end{matrix}\right.   
\end{equation}
$$ R_{d}(t)=\frac{\varepsilon _{HDV}e^{-r_{d}(t)}}{1+e^{-v_\textrm{{relative}}(x_\textrm{{AV}}-x_\textrm{{HDV}}-\sigma lv_\textrm{{relative}})}} $$
where $r_{s}(t)$\, is a coefficient of static risks at time $t$\, $v_\textrm{{relative}}$\, is a reference value to reflect the comparison of velocity of AV and each HDV, $R_{s}(t)$\, is the magnitude of dynamic risks, $\sigma$\, is the weight factor of vehicle length, $\l$\, is the vehicle length. 

A specific scenario is illustrated in Fig.3. In the depicted scenario, the blue vehicle symbolizes the AV, and the black vehicles represent the HDVs, positioned 15 and 42 meters in longitudinal. respectively. All vehicles are moving from the left to the right side of the illustration. Figure 3(b) shows that the static field, akin to the APF, radiates uniformly from the HDVs without any sharp variations, indicating a consistent risk of collision. The gradation from lighter to darker shades represents an increment in collision risk. Notably, the static field extends more broadly along the longitudinal axis than the lateral, aligning with the vehicles' direction of travel, thus mirroring the varied risk levels and movement trajectory of the HDVs. In contrast, the dynamic fields expand non-uniformly from the HDVs, with a broader spread in the direction of travel, signifying a differentiated risk profile in various directions. The hybrid field, is an amalgamation of the dynamic and static fields, holding their collective strengths to provide a comprehensive risk assessment.

\section{Decision Network and Balanced Reward Function}
 The decision making module is mainly orientated by the decision network and the balanced reward function. The decision network generates optimal control commands among the reward function during training. The balanced reward function adjusted the ratio of historical reward and current reward along with the complexity of driving scenario. The balanced reward function enhances the balanced exploration of the control policy.
\subsection{Decision Network}
 The decision network is to generate safe commands during training. The decision network consists of two sets of image-efficient actor-critic networks that receive the risk-attention states. The control policy in the actor-critic network compares the candidate control commands and select the best one based on their relative advantages.

\subsubsection{Network Structure}

The aforementioned two actor-critic networks select actions based on the states of the AV. The actor-critic network is designed based on the inputs, which are the risk-attention states. The risk-attention states consist of the driving scenario image featuring the AV and HDVs, along with the image of the hybrid risk field. Recently, convolutional neural networks demonstrates excellent capability of image classification in various domain~\cite{bharadiya2023convolutional}~\cite{bala2023monkeynet}. Therefore, two convolutional layers are used to extract essential information from the input images. Additionally, the channel attention layer and the spatial attention layer are utilized to distill crucial information pertaining to collision risks. The control policy selects control commands for collision avoidance. 
\begin{figure*}[ht]%[h]%[!p]
    \centering
    \includegraphics[width=0.9\linewidth]{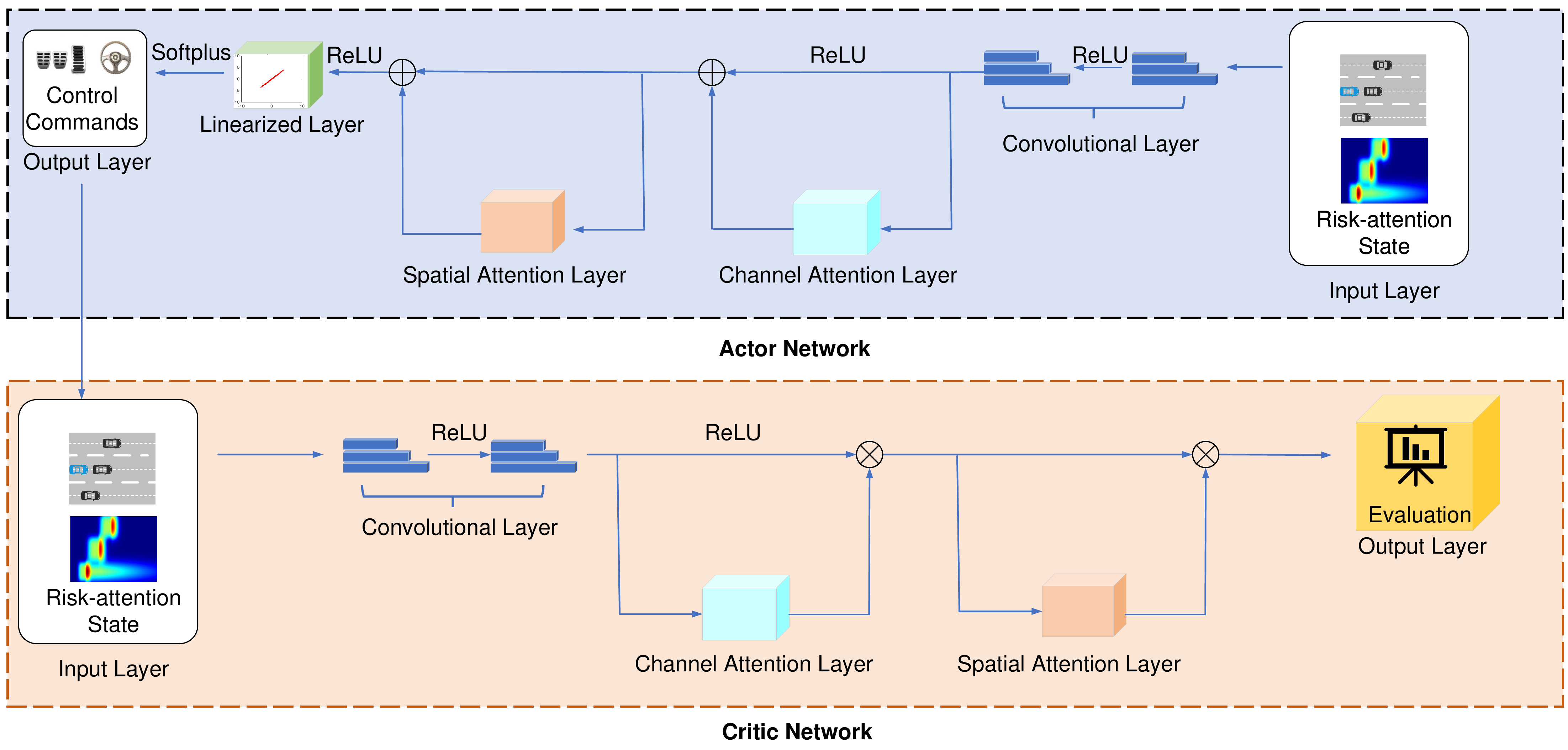}
    \caption{Structure of the image-efficient actor-critic network.}
    \label{fig2}
\end{figure*}

The evaluation of control commands is to estimate the relative advantages of the control commands. The relative advantages are served as a reference to update the actor-critic network. The actor-critic network consists of the actor network (AN) and the critic network (CN). The AN and CN have similar structures, while the AN is to generate candidate control commands and the CN is to evaluate the relative advantage for each candidate control command. 

The actor-critic network structure is demonstrated in Fig. 4. The AN consists of an input layer, two convolutional layers, a channel attention layer, a spatial attention layer, a linear layer, and an output layer. The input layer receives the current state of the racing car. The convolutional layers extract major features from the driving states at each step. The channel attention layer enhances feature representation learning. The spatial attention layer captures the subtle visual information and spatial dependencies of features. $\otimes$\, is the element-wise multiplication. The linear layer adds linearity to the actor-critic network during the training. The output layer generates the control commands. The rectified linear unit (ReLU) layer applies the  ReLU activation function to the output of the preceding layer. The \textrm{ReLU} function implements the operation \textrm{max}($x$\,, 0) on each input tensor element, where $x$\, represents the input element. The objective of ReLU layer is to introduce non-linearity into the actor-critic network during training. The expression format of convolution layer $CL$\, is expressed as 
\begin{equation}
 CL = (A,B,C)   
\end{equation}
\begin{equation}
    CL = (A,B,C)
\end{equation}

\begin{algorithm}[t]
\caption{Actor--Critic with GAE}
\LinesNumbered
\SetKwInOut{KwIn}{Input}\SetKwInOut{KwOut}{Output}
\KwIn{$S_{t+1}$, $r_t$}
\KwOut{Updated policy $\pi_{\theta}$}

\ForEach{driving sequence}{
    \For{$t=1$ \KwTo $T$}{
        Run \textbf{AN} to sample action $a_t \sim \pi_{\theta}(\cdot|s_t)$\;
        Execute $a_t$, observe $r_t$ and $s_{t+1}$; store $(s_t,a_t,r_t,s_{t+1})$ in buffer\;
    }
    Compute returns $G_t$ and advantages $A_t$ (e.g., GAE)\;
    \tcp{GAE: $A_t=\sum_{k=0}^{T-t}(\gamma\lambda)^k\,\delta_{t+k}$, 
    where $\delta_t=r_t+\gamma V_{\phi}(s_{t+1})-V_{\phi}(s_t)$}
    Update $\pi_{\theta}$ by maximizing the PPO clipped objective:
    \[
      L_{\text{clip}}=\mathbb{E}_t\big[\min(r_t(\theta)A_t,\;\mathrm{clip}(r_t(\theta),1-\epsilon,1+\epsilon)A_t)\big],
    \]
    where $r_t(\theta)=\frac{\pi_{\theta}(a_t|s_t)}{\pi_{\theta_{\text{old}}}(a_t|s_t)}$\;
    Update critic $V_{\phi}$ by minimizing $\,\mathbb{E}_t[(G_t-V_{\phi}(s_t))^2]$\;
}
\end{algorithm}

where $A$\,, $B$\,, and $C$\, indicate the number of input channels, the number of output channels, and the kernel size, respectively. 

The CN is composed of an input layer, two convolutional layers, a channel attention layer, a spatial attention layer, and an output layer. The input layer takes the output from AN and the current state of the car as inputs. The convolutional layers efficiently extract features from the driving states and the corresponding control commands at each step. The output layer selects the best control commands based on their evaluation under the current state. The best control commands are then sent to the actuators. To generate a convincing assessment, the relative advantage aims to compute long-term advantages and spans a time period $T$\,. Hence, the CN compares the performance of selected control commands with the average performance from the starting point $t$\, to $t+T$\,. The running process of actor-critic network is elaborated in Algorithm 1.

\begin{figure*}[t]%[h]%[!p]
    \centering
    \includegraphics[width=0.9\linewidth]{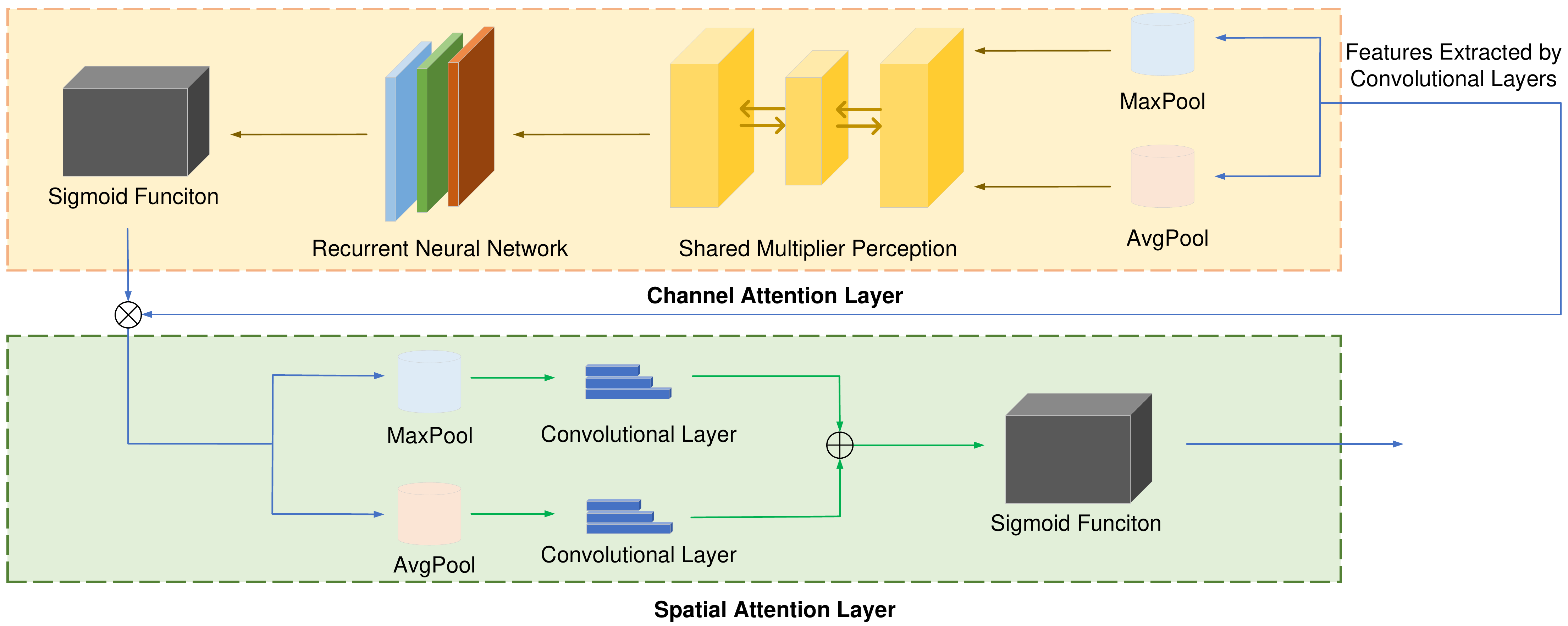}
    \caption{Structure of the channel and spatial networks.}
    \label{fig2}
\end{figure*}
The structures of designed attention network is demonstrated in Fig. 5. The channel attention network consists of a MaxPool, an AvgPool, a shared multiplier perception, a recurrent neural network and a $\textrm{sigmoid}$\, function. The MaxPool reduces the spatial dimensions of the features. The AvgPool captures overall trends of the features by computing the average value within each pooling window. The  shared multiplier perception extracts higher-level representations from the pooled features. The recurrent neural network captures temporal dependencies of the features from the shared multiplier layers, promoting the understanding of dynamic driving scenarios. The $\textrm{sigmoid}$\, function  introduces non-linearity to the network, which allows neural networks to handle complex relationships in driving features. The spatial attention layer consists of a MaxPool, an Average Pool, two convolutional layers and a  $\textrm{sigmoid}$\, function.

%\ENDFOR
%\ENDFOR

 \subsubsection{Control Policy Update of the Decision Network}
The control policy is determined by the weights of the neurons in the decision network. Therefore, the weights of the neurons should be updated to optimize the control policy. 

During driving, three types of rewards are defined: safety reward $r_{s}$, stability reward $r_{\textrm{st}}$, and efficiency reward $r_{e}$. $r_{s}$ penalizes collisions between the AV and surrounding HDVs, encouraging the AV to leave unsafe areas through timely lane changes or by adjusting the front gap. $r_{\textrm{st}}$ relates to the number of lane changes, as fewer lane changes are preferred for a smoother trajectory. $r_{e}$ reflects the velocity of the AV, with higher velocities rewarded to promote efficiency. An additional constant reward $r_{a}$ is included to adjust the overall score.

For each time step $t$, the total instantaneous reward is given by
\begin{equation}
R_{t}   = r_{s} + r_{\textrm{st}} + r_{e} + r_{a} .
\end{equation}

Instead of updating the network only when the total reward exceeds a threshold, all collected state–action–reward trajectories over a fixed horizon are used for training, following the Proximal Policy Optimization (PPO) framework. After completing a batch of trajectories, the rewards are used to compute the return-to-go and the advantage estimates $A_t$. These advantages are then used in the PPO clipped surrogate objective to update the decision network’s weights. Safety losses from collisions directly reduce $r_s$ and thus lower the corresponding advantages, guiding the learning process to avoid such scenarios in future iterations.

\begin{figure*}[t]
\centering
    \includegraphics[width=1\linewidth]{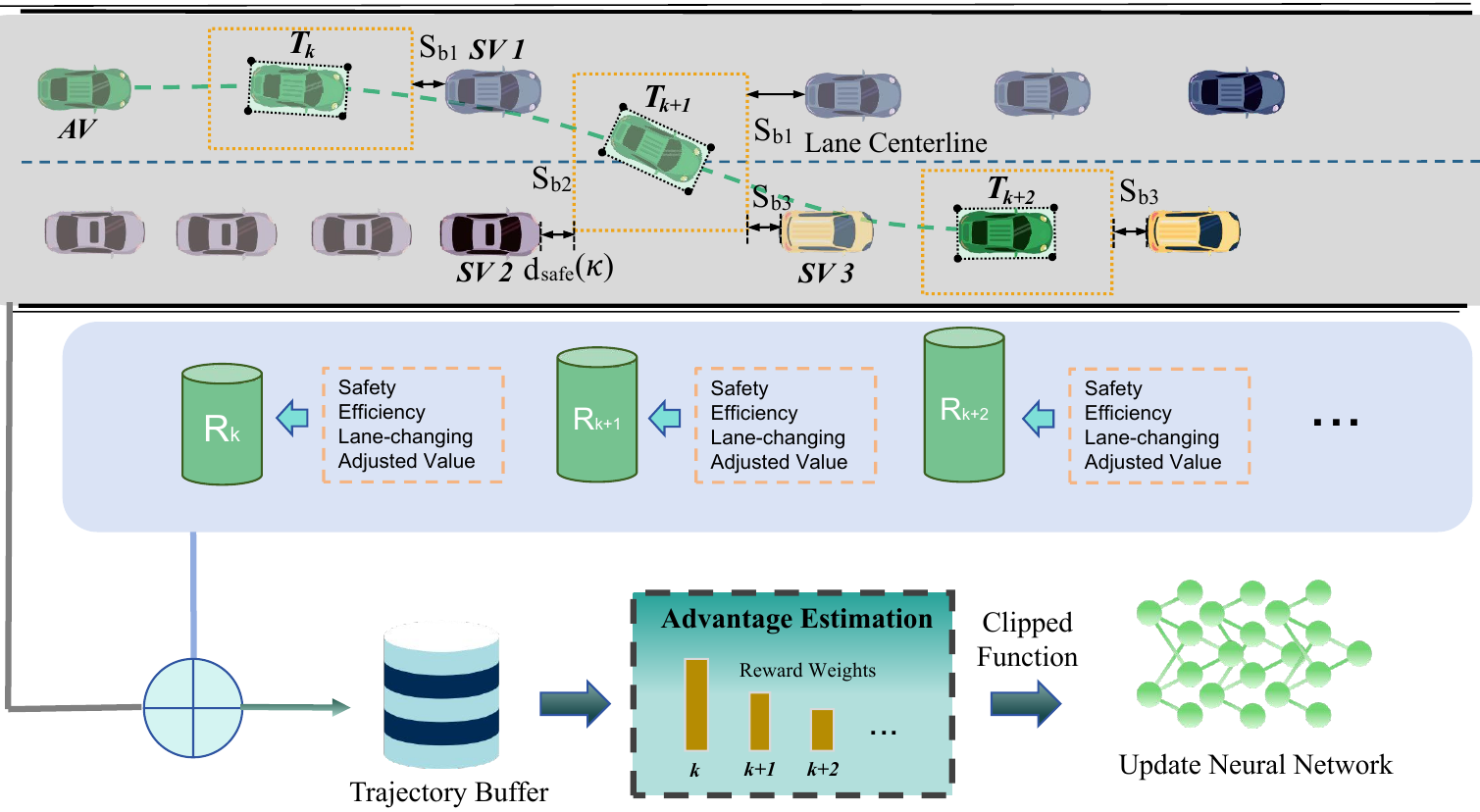}
    \caption{Control policy update process of the decision network under PPO.}
    \label{fig6}
\end{figure*}

Figure~\ref{fig6} illustrates an example of the learning process for a driving sequence in which the AV starts from the initial position and proceeds towards the goal. At the environment interaction level, the autonomous vehicle (AV) executes actions within a multi-lane highway environment that contains multiple lanes and surrounding vehicles, where the vehicle must perform efficient lane-changing maneuvers while ensuring safety constraints. The environmental state encompasses vehicle position information ($S_{b1}$, $S_{b2}$, $S_{b3}$, etc.), lane centerline information, and the motion states of surrounding vehicles, while incorporating critical parameters such as inter-vehicular safety distance constraints $d_{safe}(K)$ and temporal constraints $T_k$, $T_{k+1}$, $T_{k+2}$.

Regarding reward function design, the system adopts a multi-objective optimization strategy that decomposes the reward function into three primary dimensions: Safety, Efficiency, and Lane-changing Adjusted Value. The reward values $R_k$, $R_{k+1}$, $R_{k+2}$ corresponding to each time step $k$, $k+1$, $k+2$ comprehensively consider the trade-offs among these three dimensions, ensuring that vehicles can maintain driving safety while sustaining high traffic efficiency and possessing good lane-changing adaptability when executing lane-changing decisions.

At the learning algorithm implementation level, the system first stores the state-action-reward trajectory data generated from environment interactions in the Trajectory Buffer, subsequently computing advantage function values for each state-action pair through the Advantage Estimation module, where the advantage function reflects the superiority or inferiority of the current action relative to the average performance level. During the network update phase, the system employs the core concept of the PPO algorithm by utilizing a Clipped Function to constrain the magnitude of policy updates, preventing excessive changes in the policy network and thereby ensuring the stability of the learning process. Finally, through backpropagation algorithms, the neural network parameters are updated to achieve continuous policy optimization and improvement, enabling autonomous vehicles to learn safer, more efficient, and more intelligent lane-changing strategies.

\subsection{Balanced Reward Function}
The reward function is the feedback module that evaluates the actions selected by the decision network. In racing scenarios, velocity and collision avoidance are the two primary factors influencing performance. Velocity measures efficiency, while collision avoidance ensures safety. An effective reward function should guide the decision network toward actions that avoid collisions with HDVs and track boundaries while minimizing lap time.

However, traditional reward functions often treat all time steps equally, which causes the average reward to be dominated by earlier high-reward actions. This can mask the impact of recent poor decisions. The averaged reward is defined as
\begin{equation}
r_{\textrm{ave}}   = 0.99 \, r_{\textrm{ave}} + 0.01 \, r_{\textrm{current}},
\end{equation}
where $r_{\textrm{ave}}$ and $r_{\textrm{current}}$ represent the averaged historical reward and the reward of the current step, respectively.

Collisions with HDVs at high speeds are particularly critical and should be emphasized. Since the averaged reward cannot effectively highlight these “corner cases,” a hyperparameter $\gamma$ is introduced to balance the influence of historical and current rewards:
\begin{equation}
r  = (1-\gamma) \, r_{\textrm{ave}} + \gamma \, r_{\textrm{current}},
\end{equation}
where $\gamma$ directs the AV’s attention toward risky situations. The current reward $r_{\textrm{current}}$ is determined by the aggregated risk levels of surrounding HDVs using the hybrid risk field. When risk levels are high, the current decision is weighted more heavily.

The historical average reward is computed as
\begin{equation}
r_{\textrm{ave}} = \frac{1}{N-1} \sum_{i=1}^{N-1} s_{f}^{i},
\end{equation}
where $N$ is the current step index, and $s_f^{i}$ is the total reward at the $i^{\textrm{th}}$ step. The term $r_{\textrm{current}}$ equals $s_f$ at step $N$. This formulation promotes a safety-aware and forward-looking strategy.

To ensure stable policy improvement, PPO employs a clipped surrogate objective that constrains the update magnitude of policy parameters:
\begin{equation}
L_{\textrm{clip}} = \hat{\mathbb{E}}_t \left[ \min \left( R_t(\theta) A_t,\; \textrm{clip}(R_t(\theta),\, 1-\epsilon,\, 1+\epsilon) A_t \right) \right],
\end{equation}
where $R_t(\theta)$ is the probability ratio between the new and old policies, $\epsilon$ is a hyperparameter controlling the update range, and $A_t$ is the estimated advantage. This clipping mechanism prevents excessively large updates that could destabilize the policy, ensuring smoother and more reliable convergence.

\section{Safety-assisted Mechanism}
In this section, this paper presents the proposed safety-assisted mechanism. Specifically, at each time step t, with the generated control commands form decision network as discussed above, the safety-assisted mechanism first detects and prevents the potential collisions for the AVs. The potential collisions can be divided into the collisions during lane-changing and the collisions during lane keeping. The detection of collisions during lane changing rely on a lane-changing model. The detection of collisions during lane keeping rely on a fixed safety .

\subsection{Detection of collisions during lane changing}

Before proceeding to the actuator, the safety-assisted mechanism evaluates the risk along the entire lane-changing trajectory using the previously defined static and dynamic risk fields. Training sequences with high collision risk have no contribution to the collision-free control policy. Therefore, the safety-assisted mechanism predicts and filters out high-risk lane changes before execution, preventing the generation of useless training sequences.

\begin{figure}[t]
\centering
    \includegraphics[width=1\linewidth]{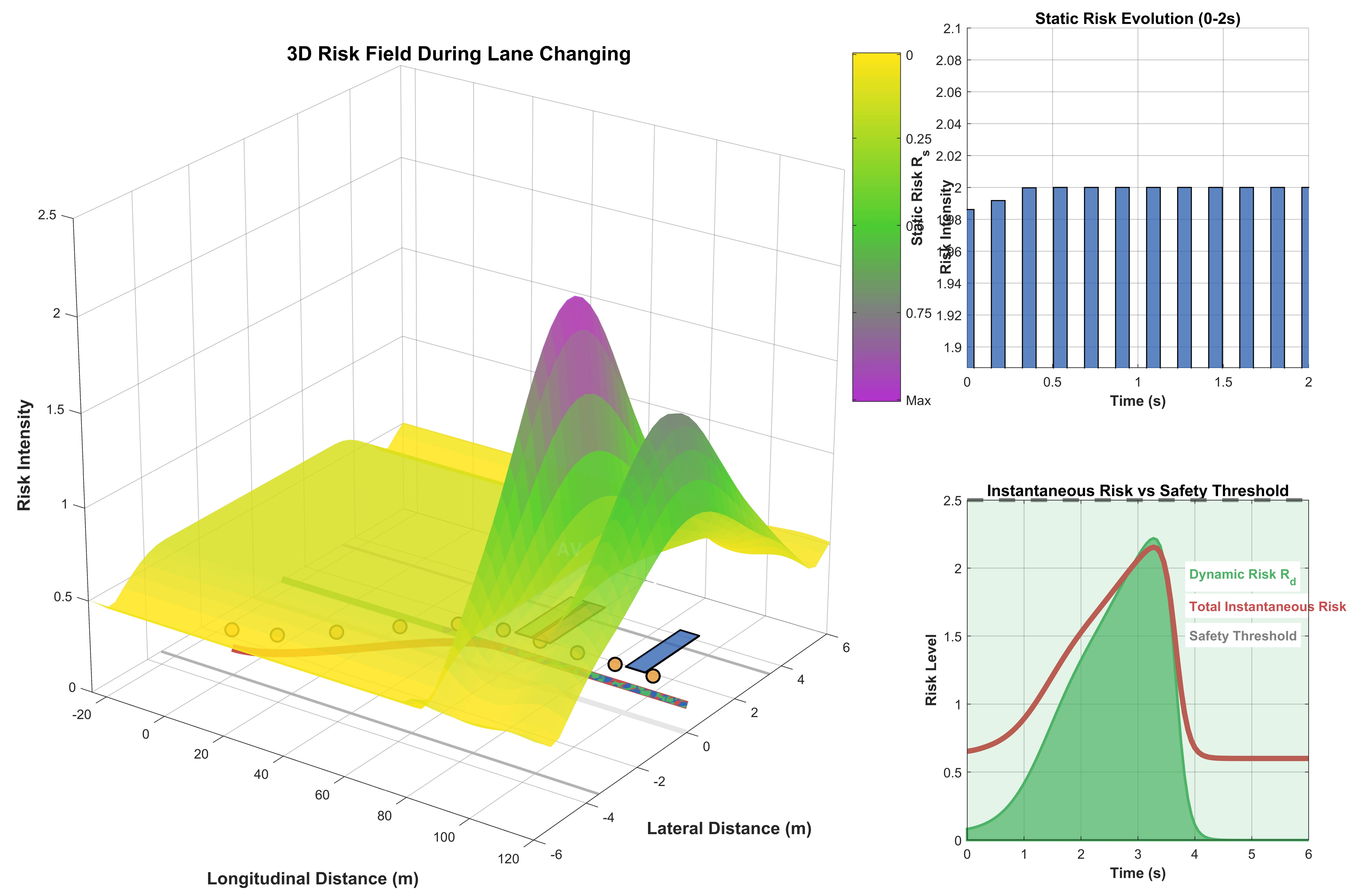}
    \caption{Safety-assisted lane-changing process}
    \label{fig7}
\end{figure}

As illustrated in Fig.~\ref{fig7}, three participants are involved in the lane-changing process: the front vehicle (FV), the autonomous vehicle (AV), and the rear vehicle (RV). The primary collision threat comes from the interaction between the AV's lane-changing trajectory and the RV's trajectory. The lane-changing trajectory of the AV is modeled as an ideal cubic polynomial curve \(y(x)\) that is smooth and consistent with human driving habits:
\begin{equation}
y(x) = -2 + \frac{2}{x_r}x - \frac{3}{x_r^2}x^2 + \frac{1}{x_r^3}x^3,
\label{eq:lc_traj_riskfield}
\end{equation}
where \(x_r\) is the longitudinal distance from the start point to the mid-point of the lane change.

Instead of computing the time-to-collision at a single potential collision point, we discretize the lane-changing trajectory into \(N\) sampling points \(\{(x(t_i), y(t_i))\}_{i=1}^N\) and evaluate the instantaneous risk at each point based on the static risk field:
\begin{align}
R_s(t_i) &= \varepsilon_{\text{obs}} e^{-r_s(t_i)}, \\
r_s(t_i) &= \left(\frac{\Delta x_i}{\xi_x}\right)^{2\rho} + \left(\frac{\Delta y_i}{\xi_y}\right)^{2\rho},
\end{align}
and the dynamic risk field:
\begin{align}
R_d(t_i) &= \frac{\varepsilon_{\text{HDV}} e^{-r_d(t_i)}}{1+e^{-v_{\text{relative}}\left(\Delta x_i - \sigma l v_{\text{relative}}\right)}}, \\
r_d(t_i) &= \left(\frac{\Delta x_i}{\xi_v}\right)^{2\lambda} + \left(\frac{\Delta y_i}{\xi_y}\right)^{2\lambda},
\end{align}
where \(\Delta x_i = x_{\text{AV}}(t_i) - x_{\text{HDV}}(t_i)\) and \(\Delta y_i = y_{\text{AV}}(t_i) - y_{\text{HDV}}(t_i)\) are the relative positions, and \(v_{\text{relative}}\) is defined as in the dynamic risk field formulation.

The cumulative lane-changing risk is then obtained as:
\begin{equation}
R_{\text{total}} = \sum_{i=1}^N \left[ w_s R_s(t_i) + w_d R_d(t_i) \right],
\label{eq:total_risk_lanechange}
\end{equation}
where \(w_s\) and \(w_d\) are weighting coefficients for static and dynamic risk contributions.
Finally, the lane change is considered safe if:
\begin{equation}
R_{\text{total}} \le R_{\text{safe}},
\label{eq:lanechange_safe_condition}
\end{equation}
where \(R_{\text{safe}}\) is a predefined safety threshold. If the condition in (\ref{eq:lanechange_safe_condition}) is not met, the lane change is discarded before policy execution.

Figure~\ref{fig7} presents a comprehensive 3D visualization of the collision detection mechanism during lane-changing maneuvers, demonstrating the spatial and temporal evolution of risk fields in a multi-vehicle scenario. The main 3D surface plot illustrates the risk distribution across the longitudinal-lateral plane, where the blue-to-purple gradient indicates increasing risk intensity, with the highest risk concentrations occurring around the rear vehicle (RV) position due to the critical interaction between the autonomous vehicle's lane-changing trajectory and the approaching RV. The autonomous vehicle (AV) trajectory, represented by the red curve, exhibits a smooth cubic polynomial transition from the bottom lane to the middle lane, strategically avoiding the highest-risk regions while maintaining vehicle dynamics constraints.

The static risk evolution subplot reveals minimal variation in boundary-related risks, fluctuating between 1.986 and 2.0, which demonstrates the effectiveness of the lane-changing trajectory design in maintaining safe distances from road boundaries throughout the maneuver. More significantly, the risk comparison analysis shows distinct behavioral patterns for different risk components: the dynamic risk peaks at approximately $t = 2.5$ seconds with a maximum value of 2.2, corresponding to the moment of closest proximity between the AV and RV, followed by a rapid decrease as the vehicles separate. The instantaneous total risk closely follows the dynamic risk pattern, confirming that inter-vehicle interactions dominate the overall risk profile during lane-changing operations.

The cumulative risk analysis provides crucial insights into the integrated safety assessment, showing a monotonic increase that stabilizes at approximately 2.5, which precisely aligns with the predefined safety threshold. This near-threshold convergence indicates a marginally safe lane-changing scenario where the maneuver is deemed acceptable but operates at the safety boundary. The risk field sampling points distributed along the AV trajectory enable discrete risk evaluation at multiple positions, ensuring comprehensive coverage of the entire lane-changing path. The safety assessment mechanism successfully identifies this as a borderline safe maneuver, where the maximum instantaneous risk remains below the threshold while the cumulative risk approaches but does not exceed the safety limit, demonstrating the system's capability to distinguish between acceptable and unacceptable lane-changing scenarios while maintaining practical operational flexibility in dynamic traffic environments.

\section{Simulation Results}
The proposed RIBPPO-S is evaluated on a three-lane highway considering convergence, final performance, collision rate, and AV speed variation. To assess generalizability, two traffic-flow settings with 4--6 surrounding HDVs are tested. The HDV selection follows: (1) the front and rear vehicles in the AV’s lane are always included; (2) for adjacent lanes, if the total gap between the AV and nearest front/rear vehicles in its lane exceeds $20~\text{m}$, all vehicles within the AV’s longitudinal range or with lateral overlap are counted; otherwise, only the nearest vehicle per adjacent lane is included. 
\begin{figure*}[t]
    \centering
    \subfigure[]{
    \label{fig:blockloads-torque}
    \includegraphics[width=.3\textwidth, height=90pt]{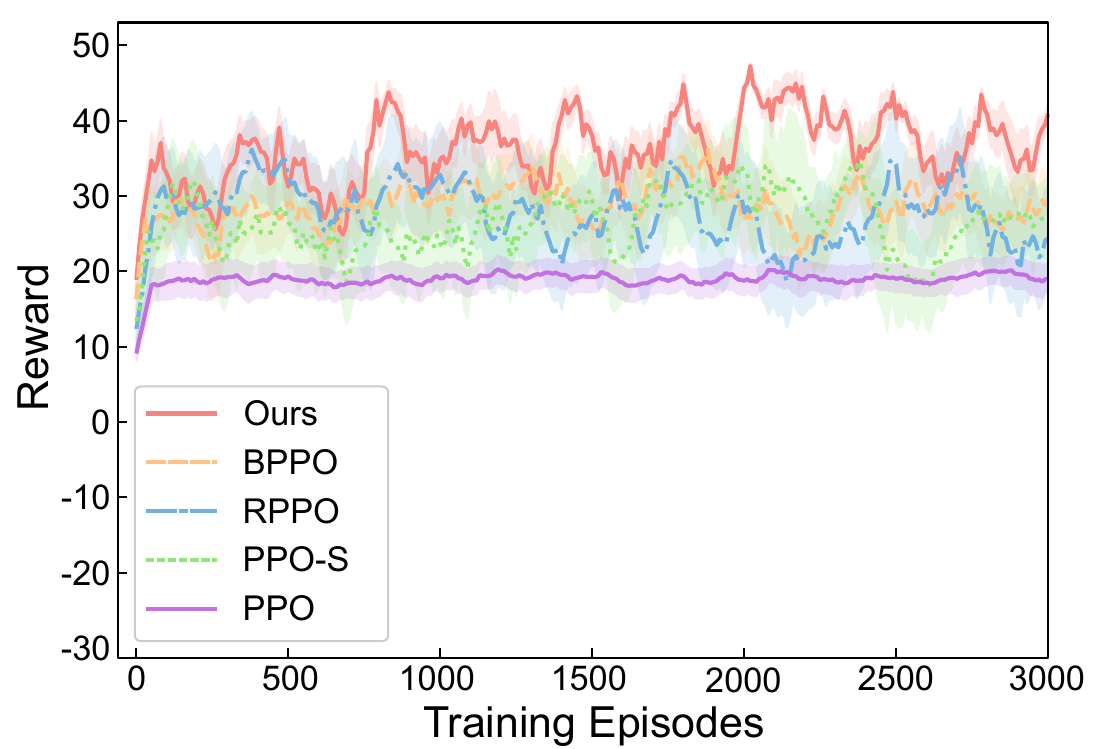}}  
    \centering
    \subfigure[]{
    \label{fig:blockloads-MAP}
    \includegraphics[width=.3\textwidth, height=90pt]{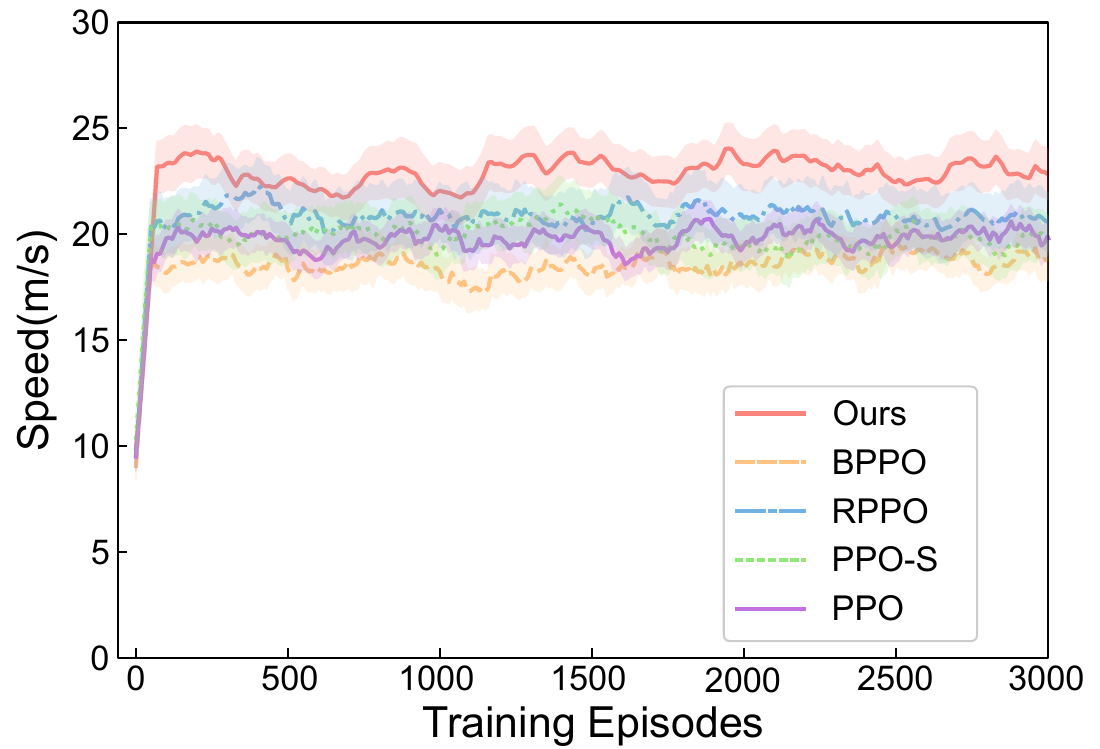}}
    \centering
    \subfigure[]{
    \label{fig:blockloads-speed}
    \includegraphics[width=.3\textwidth, height=90pt]{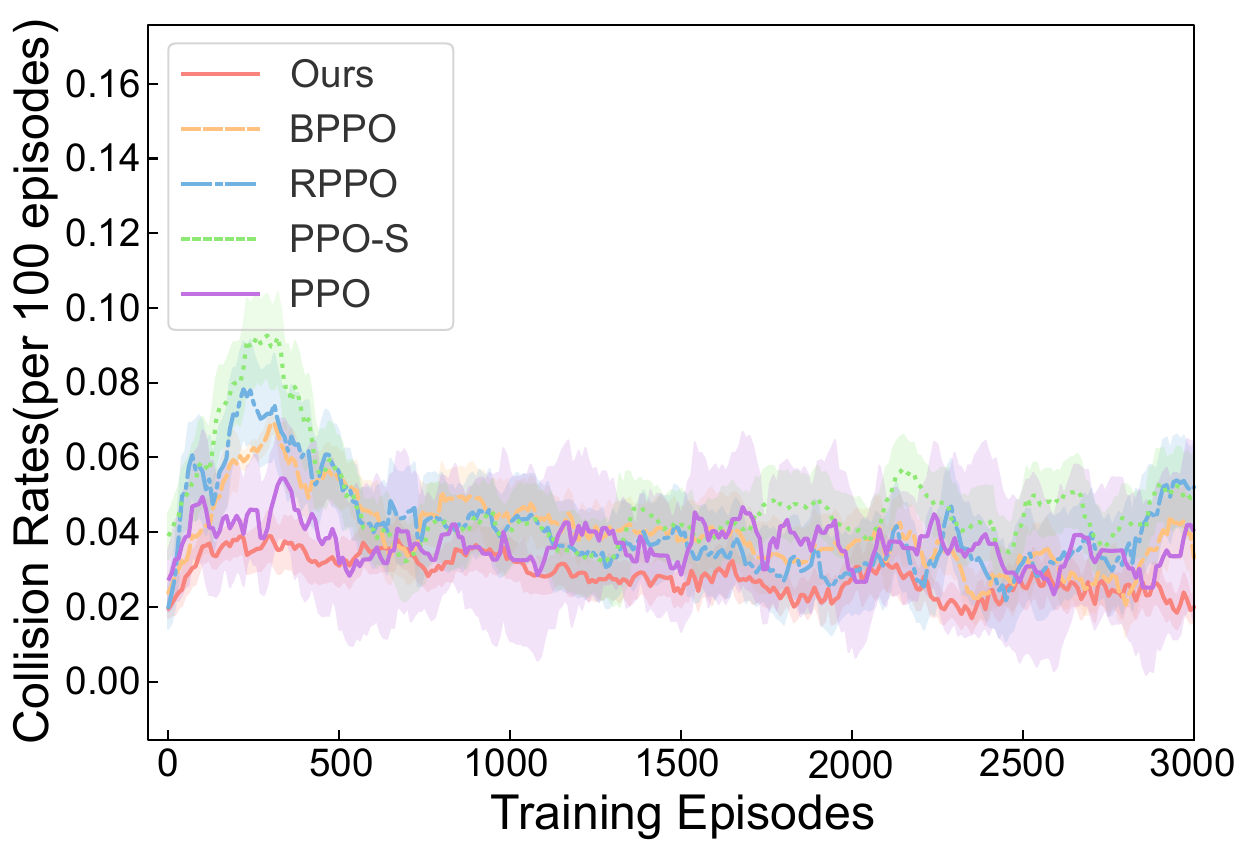}}
    
\caption{Performance of the RIBPPO-S, BPPO, RPPO, PPO-S, and PPO during the converging. (a) rewards; (b) speed variations; (c) collision rates.}
    \label{fin}
\end{figure*}

\begin{figure}[t]%[!p]
    \centering
    \includegraphics[width=0.55\linewidth]{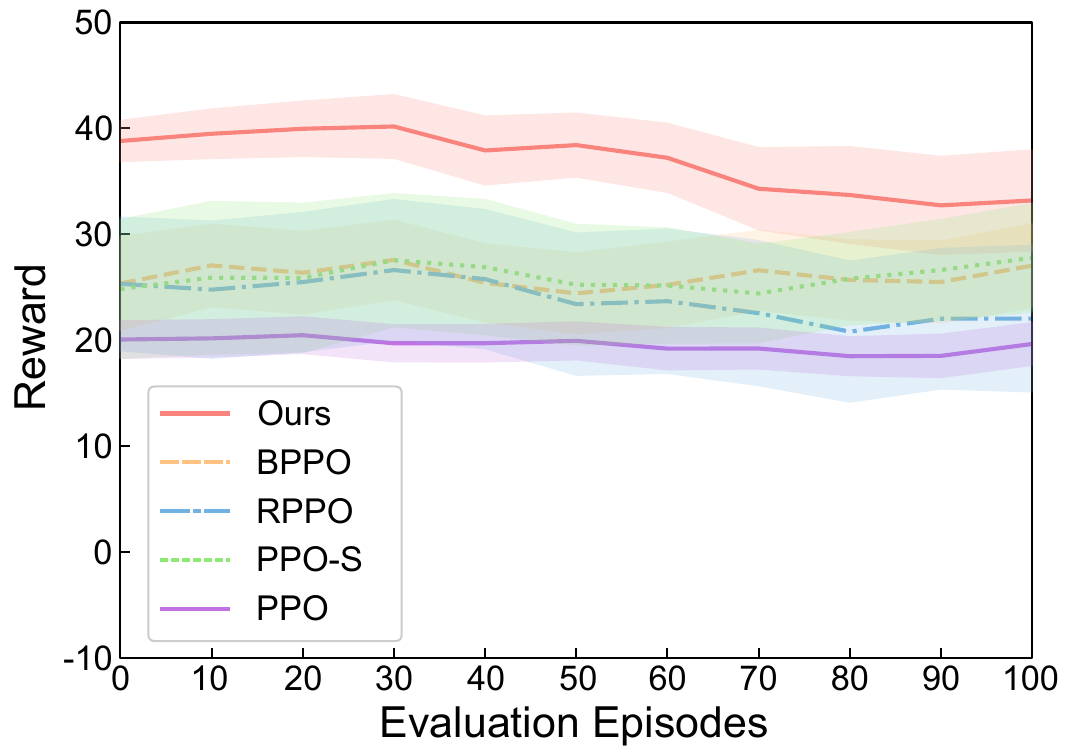}
    \caption{Reward of the RIBPPO-S, BPPO, RPPO, PPO-S, and PPO after the converging.}
    \label{figaa}
\end{figure}

Two verification studies are conducted: \textit{(i)} RIBPPO-S is compared with PPO, BPPO, RPPO, and PPO-S; \textit{(ii)} RIBPPO-S is compared with DDPG~\cite{basile2022ddpg}, PPO~\cite{ye2020automated}, A2C~\cite{zhai2023model}, and DQN~\cite{wang2018automated} under both traffic settings.

Experiments use Highway-env~\cite{highway-env} with three $4~\text{m}$-wide lanes. The AV’s desired speed is randomly set between $23$--$25~\text{m/s}$ (max $40~\text{m/s}$) and incorporated into the MPC objective. HDV positions and speeds are randomized within small ranges. Implementation is based on Python~3.6, PyTorch~1.10.0, Ubuntu~20.04.6~LTS, an Intel\textsuperscript{\textregistered} Core\texttrademark\ i5-12600KF (12th Gen, 16 threads), NVIDIA RTX~3090 GPU, and $64~\text{GB}$ RAM.

\subsection{Comparison with PPO, RPPO, BPPO, and PPO-S}
Fig. 8 presents the convergence, driving speed variation, collision rates of the proposed RIBPPO-S and RPPO, BPPO, PPO-S, and PPO during the converging, respectively. Fig. 10 presents the rewards of the proposed RIBPPO-S and four other algorithms using converged policy. The other four algorithms include PPO, BPPO, RPPO, and PPO-CS. As shown in Fig. 8(a), the proposed RIBPPO-S reaches a higher reward and faster convergence compared to four other algorithms. The rewards of the proposed RIBPPO-S, RPPO, BPPO, PPO-S and PPO, are around 39, 31, 25, 22, and 18, respectively. As shown in Fig. 8(b), the proposed RIBPPO-S achieves a higher average speed and faster convergence compared to four other algorithms. The average speed and variance of the RIBPPO-S, RPPO, PPO-S, PPO, and BPPO are (23, 3), (21.5, 2), (20, 3), (19, 2), and (17.5, 2.5), respectively. As shown in Fig. 7(c), the proposed RIBPPO-S has a lower collision rate and faster convergence. The collision rate of the proposed RIBPPO-S, PPO, RPPO, BPPO, and PPO-S are below 0.04 per 100 episodes, 0.06 per 100 episodes, 0.08 per 100 episodes, 0.075 per 100 episodes, and  0.1 per 100 episodes, respectively. As shown in Fig. 8, the proposed RIBPPO-S has a higher reward after the converging compared to four other algorithms. The rewards of the proposed RIBPPO-S, PPO-S, BPPO, RPPO, and PPO, are around 35, 27, 25, 22, and 20, respectively. The above analysis demonstrates the effectiveness in terms of the safety and efficiency.
\begin{figure*}[t]
    \centering
    \subfigure[]{
    \label{fig:blockloads-torque}
    \includegraphics[width=.3\textwidth, height=90pt]{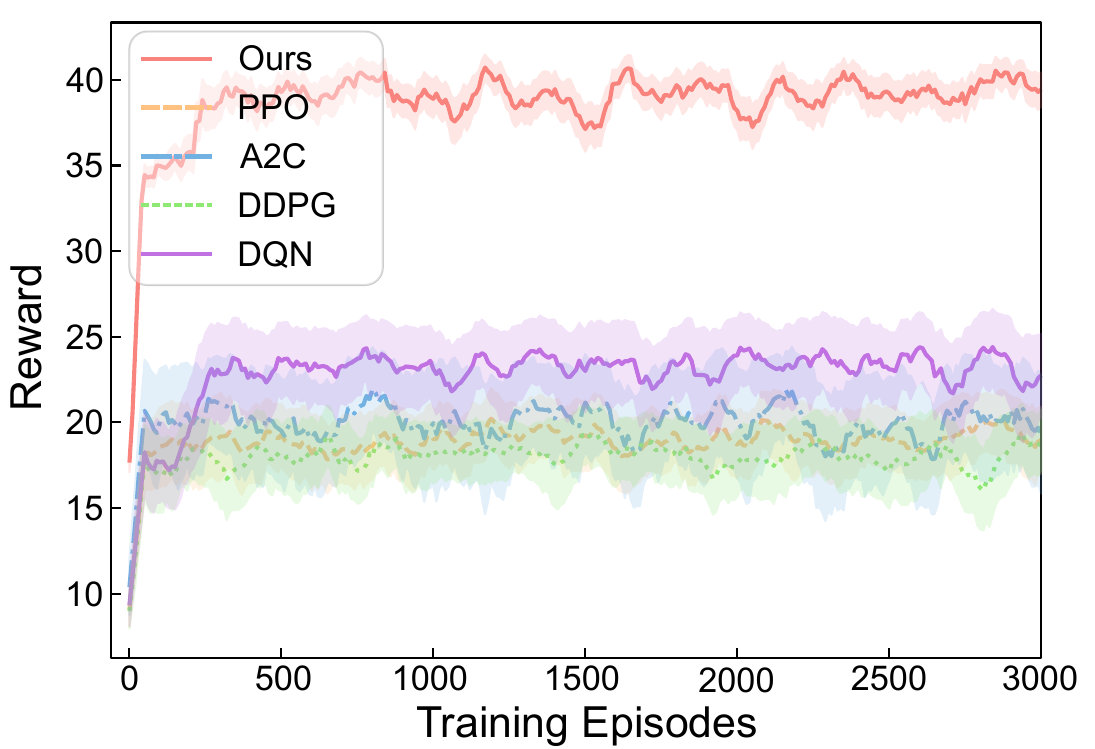}}  
    \centering
    \subfigure[]{
    \label{fig:blockloads-MAP}
    \includegraphics[width=.3\textwidth, height=90pt]{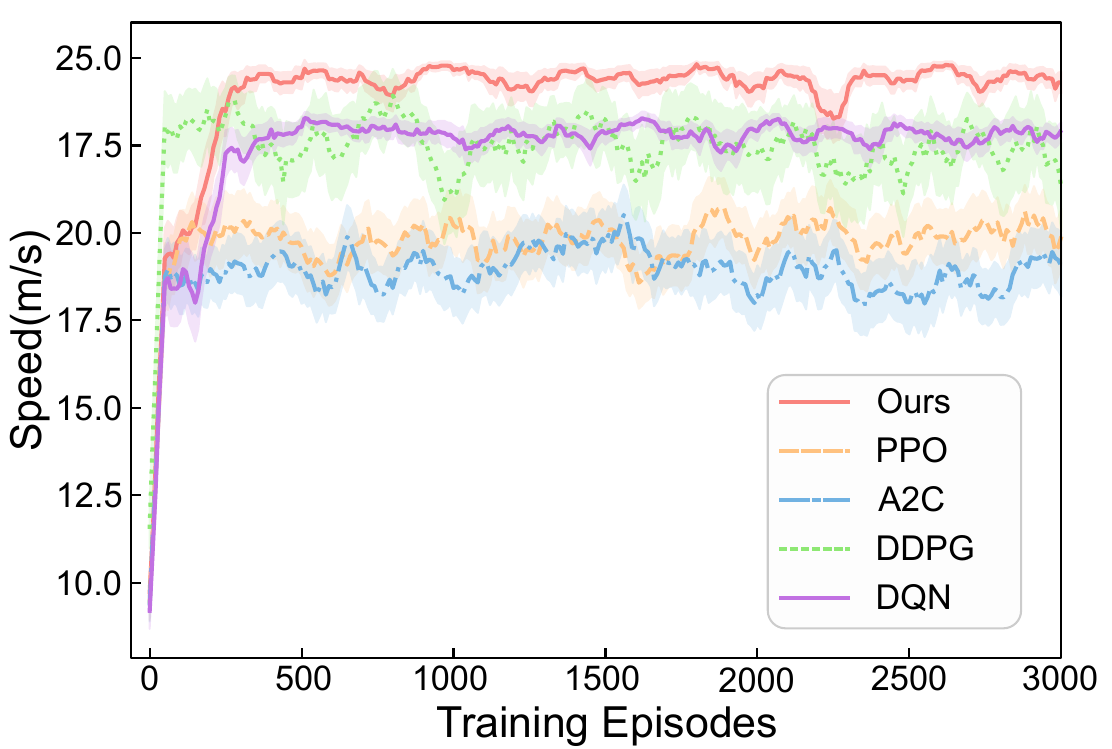}}
    \centering
    \subfigure[]{
    \label{fig:blockloads-speed}
    \includegraphics[width=.3\textwidth, height=90pt]{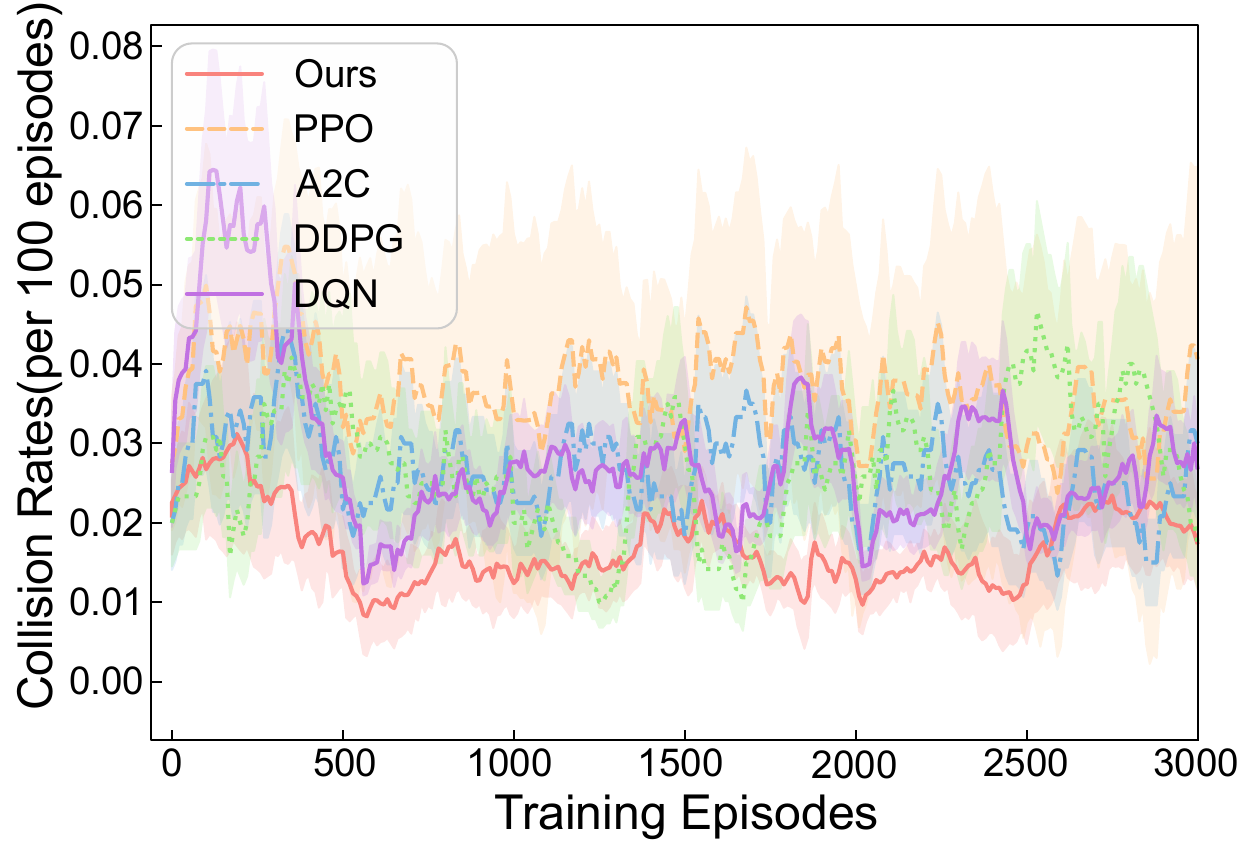}}
    
\caption{Performance of the RIBPPO-S, PPO, A2C, DDPG and DQN during the converging in the normal traffic mode. (a) rewards; (b) speed variations; (c) collision rates.}
    \label{fin}
\end{figure*}

\begin{figure}[t]%[!p]
    \centering
    \includegraphics[width=0.5\linewidth]{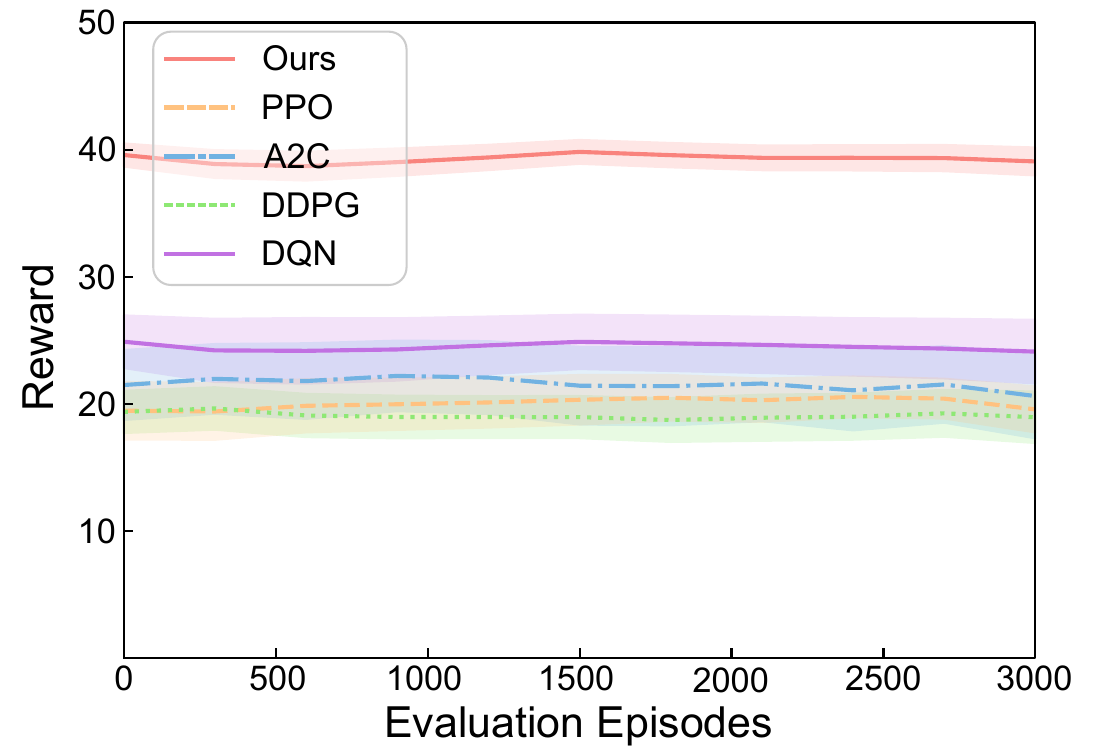}
    \caption{Rewards of the RIBPPO-S, PPO, A2C, DDPG, and DQN in the normal traffic mode after the converging.}
    \label{figaa}
\end{figure}

\subsection{Comparison with other popular DRL benchmarks}
 As shown in Fig. 10(a), the proposed RIBPPO-S reaches a higher reward and faster convergence compared to the four benchmark DRL algorithms. The rewards of the proposed RIBPPO-CS, DQN, A2C, PPO and DDPG, are around 39, 23, 20, 18, and 17, respectively. As shown in Fig. 10(b), the proposed RIBPPO-S achieves a higher average speed and faster convergence compared to the four benchmark DRL algorithms. The average speed and variance of the RIBPPO-S, DQN, DDPG, PPO, and A2C are around (23, 3), (22.5, 1.5), (22, 5), (20, 3), and (19, 3), respectively. As shown in Fig. 10(c), the proposed RIBPPO-S has a lower collision rate and faster convergence. The collision rates of the proposed RIBPPO-S, A2C, DDPG, PPO, and DQN are below 0.04 per 100 episodes, 0.045 per 100 episodes, 0.05 per 100 episodes, 0.06 per 100 episodes, and 0.065 per 100 episodes, respectively. As shown in Fig. 11, the proposed RIBPPO-S has a higher reward after the converging compared to the four benchmark DRL algorithms. The rewards of the proposed RIBPPO-S, DQN, A2C, PPO and DDPG, are around 38, 23, 20, 19, and 18, respectively.

\subsection{Examples in the 100-$\textrm{m}$ and 200-$\textrm{m}$ Long-Distance Scenarios}
\begin{figure}[t]%[!p]
    \centering
    \includegraphics[width=0.8\linewidth]{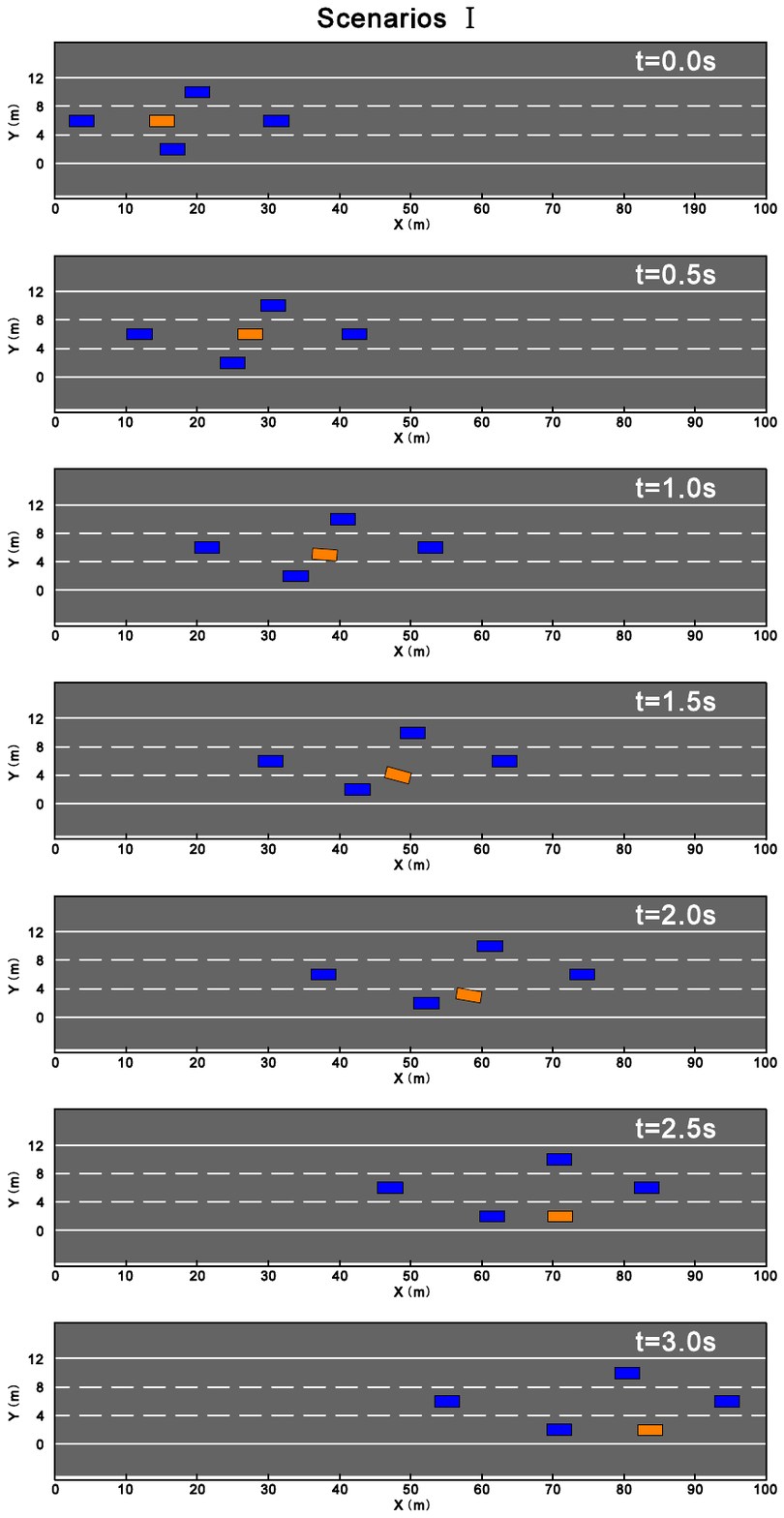}
    \caption{Example illustration in 100-$\textrm{m}$ Long-Distance Scenario.}
    \label{figaa}
\end{figure}
\begin{figure}[h]%[!p]
    \centering
    \includegraphics[width=0.8\linewidth]{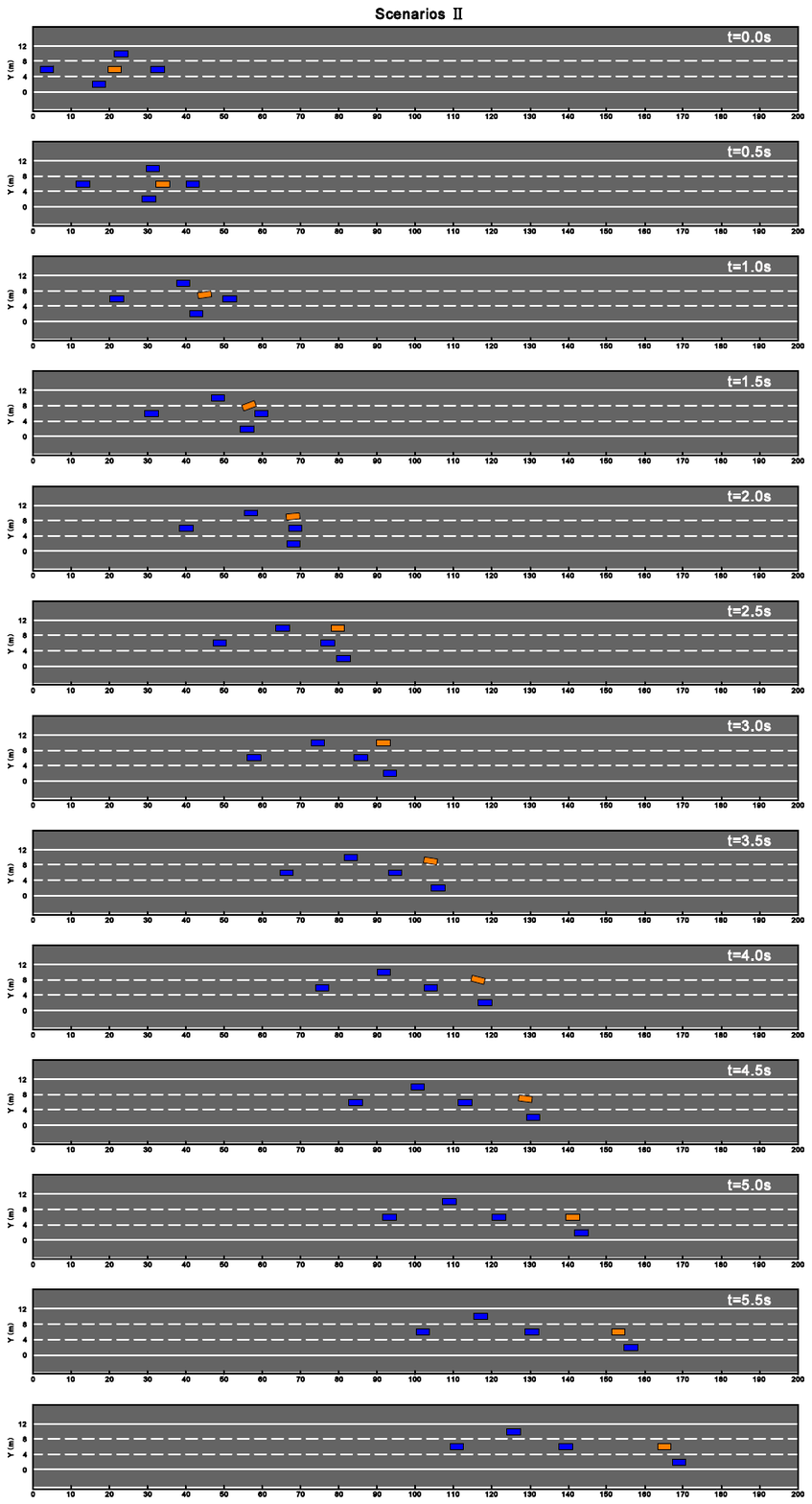}
    \caption{Example illustration in 200-$\textrm{m}$ Long-Distance Scenario.}
    \label{figaa}
\end{figure}
Using the proposed RIBPPO-S, the AV successfully performs lane changes in both scenarios. Figs.~12 and 13 illustrate the AV's interactions with surrounding HDVs between $0$--$3~\text{s}$ in the 100-$\textrm{m}$ and 200-$\textrm{m}$ scenarios, respectively. The AV and HDVs are depicted as orange and blue blocks. Seven time instants are shown in the 100-$\textrm{m}$ scenario: $t = 0~\text{s}$, $0.5~\text{s}$, $1.0~\text{s}$, $1.5~\text{s}$, $2.0~\text{s}$, $2.5~\text{s}$, and $3.0~\text{s}$. The three lanes are referred to as the \emph{upper}, \emph{middle}, and \emph{bottom} lanes from top to bottom. 

In the \textbf{100-$\textrm{m}$} scenario, the AV initiates its first lane change from the middle to the bottom lane at $t = 1.0~\text{s}$, completing it by $t = 2.5~\text{s}$. During other times, the AV maintains stable driving on the highway.

In the \textbf{200-$\textrm{m}$} scenario, the AV begins the first lane change from the middle to the upper lane at $t = 1.0~\text{s}$, completing it by $t = 2.5~\text{s}$. The second lane change, from the middle to the upper lane, starts at $t = 3.5~\text{s}$ and finishes at $t = 5.0~\text{s}$.

In both scenarios, the AV maintains stable and safe motion throughout, aside from the necessary lateral maneuvers.

%%%%%%%%%%%%%%%%%%%%%%%%%%%%%%%%%%%%%%%%%%%%%%%%%%%%%%%%%%%%%%%%%%%%%%%%%%%%%%%%%%%%%%%%%%%%%%%%%%%%%%%%%%%%

\section{Conclusion}
This paper presents a top-down safe planner to improve the safety and lane-changing efficiency of AVs on the ramp, considering the uncertainty of prediction and driving styles of surrounding human vehicles. 
The planner incorporates a prediction error ellipse to suggest dangerous areas, a data-based grading system to distinguish personal driving styles, a collision detection method considering TTR to make driving safer, and an IDM model to ensure safety of the following process.
The planner is tested considering conservative HDVs, moderate HDVs, aggressive HDVs and mixed platoons. The results demonstrate that the AVs with the proposed planner can drive with no collisions, with relatively high velocity and low THW for a long period. Comparisons show that the proposed planner outperforms the existing methods in terms of shorter lane changing time, higher average velocities,  and fewer collisions to the dangerous areas. Future research will be testing the proposed planner under challenging conditions such as icy roads and curvy roads.

\bibliographystyle{IEEEtran}
\bibliography{IEEEabrv,zq_lib}

\vfill

\end{document}